\documentclass[conference]{IEEEtran}
\IEEEoverridecommandlockouts
\usepackage{cite}
\usepackage{algorithmic}
\usepackage{graphicx}
\usepackage{textcomp}
\usepackage{xcolor}
\usepackage{tikz}
\usepackage[utf8x]{inputenc}
\usepackage{listings,multicol}

\usepackage{subfig}
\usepackage{siunitx}
\DeclareSIUnit\pixel{px}
\usepackage{tabularx}
\usepackage{float}
\usepackage{url}

\usepackage{amssymb, amsmath, amsfonts, xfrac}

\newcommand{\airlab}{AIRLab}
\newcommand{\fixed}{{I_{\text{F}}}}
\newcommand{\moving}{\ensuremath{I_{\text{M}}}}
\def\R{{\rm I\!R}}

\DeclareMathOperator*{\argmin}{arg\,min}

\newcommand{\air}[1]{%
\begin{tikzpicture}[#1]%
    \draw (0, 0)
    -- ++ (-60:1.5ex)
    -- ++ (180:1.5ex)
    -- ++ (60:1.5ex)
    -- cycle;
    \draw (0, 0)
       ++ (270:0.5ex)
       ++ (0:0.75ex)
    -- ++ (180:1.5ex);
\end{tikzpicture}%
}

\DeclareUnicodeCharacter{128769}{\air{}}

\def\BibTeX{{\rm B\kern-.05em{\sc i\kern-.025em b}\kern-.08em
    T\kern-.1667em\lower.7ex\hbox{E}\kern-.125emX}}
    
\usepackage{filecontents}

\begin{document}
\lstset{language=Python, basicstyle=\ttfamily\scriptsize}

\title{\airlab: Autograd Image Registration Laboratory}

\author{\IEEEauthorblockN{Robin Sandk{\"u}hler, Christoph Jud, Simon Andermatt, 
                           and Philippe C. Cattin}
\IEEEauthorblockA{Department of Biomedical Engineering\\
University of Basel \\
 Allschwil, Switzerland \\
\small{\texttt{\{robin.sandkuehler, christoph.jud, simon.andermatt, philippe.cattin\}@unibas.ch}}
}}

\maketitle 

\begin{abstract}
Medical image registration is an active research topic and forms a basis
for many medical image analysis tasks. Although image registration is a rather
general concept specialized methods are usually required to target a specific registration problem.
The development and implementation of such methods has been tough so far as
the gradient of the objective has to be computed. Also, its evaluation has to be
performed preferably on a GPU for larger images and for more complex transformation models and regularization terms.
This hinders researchers from rapid prototyping and poses hurdles to reproduce
research results. There is a clear need for an environment which hides this complexity
to put the modeling and the experimental exploration of registration methods into the foreground.
With the ``Autograd Image Registration Laboratory'' (\airlab), we introduce an open laboratory
for image registration tasks, where the analytic gradients of the objective function
are computed automatically and the
device where the computations are performed, on a CPU or a GPU, is transparent. It is meant 
as a laboratory for researchers and developers enabling them to rapidly try out new ideas for
registering images and to reproduce registration results which have already been published.
\airlab{} is implemented in Python using PyTorch as tensor and optimization library and SimpleITK
for basic image IO. Therefore, it profits from recent advances made by the machine learning
community concerning optimization and deep neural network models. 

The presented draft of this paper outlines \airlab{} with first code snippets and performance
analyses. A more exhaustive introduction will follow as a final version soon.
\end{abstract} 


\section{Introduction}
The registration of images is a growing research topic and forms an integral part in many
medical image analysis tasks \cite{Viergever2016}. It is referred to as 
the process of finding corresponding structures within different images. There is a large number of
applications where image registration is inevitable such as e.g.~the fusion of different modalities,
monitoring anatomical changes, population modelling or motion extraction.

Image registration
is a nonlinear, ill-posed problem which is approached by optimizing a regularized objective.
What is defined as quite general requires usually specialized objective functions and implementations
for applying it to specific registration tasks. The development of such specific registration methods
has been tough so far and their implementation tedious. This is because gradients have to be computed
within the optimization whose implementations are error-prone, especially for 3D objectives. Furthermore,
for large 3D images, the computational demand is usually high and a parallel execution on a GPU unavoidable.
These are problems which hinder researchers from playing around with different combinations of
objectives and regularizers and rapidly trying out new ideas. Similarly, the effort to reproduce registration
results is often out of proportion. There is a clear need for an environment which hides this complexity,
enables rapid prototyping and simplifies reproduction.
\begin{figure}
\centering
 \subfloat[Fixed]{%
 \raisebox{0mm}{
    \includegraphics[width=0.2\textwidth]{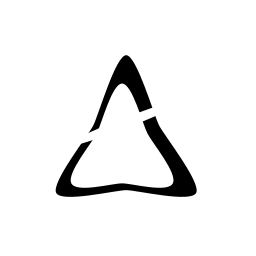}
    }
 }
\subfloat[Moving]{%
\raisebox{0mm}{
   \includegraphics[width=0.2\textwidth]{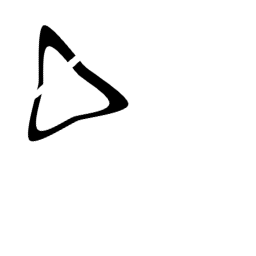}
   }
}\\
\subfloat[Warped]{%
\raisebox{0mm}{
   \includegraphics[width=0.2\textwidth]{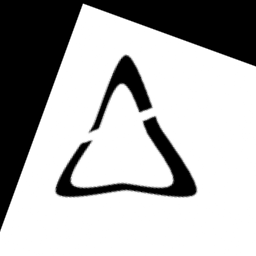}
   }
}

\caption{(a) Fixed \airlab{} image, (b) moving \airlab{} image and (c) warped moving \airlab{} image after registration.}
\label{fig:rigid_example}
\end{figure}
In this paper, we introduce ``Autograd Image Registration Laboratory'' (\airlab), an image registration
environment - or a laboratory - for rapid prototyping and reproduction of image registration methods. 
Thus, it addresses researchers and developers and simplifies their work in the exploration of different 
registration methods, in particular also with upcoming complex deep learning approaches.  
It is written in Python and based on the tensor library PyTorch \cite{paszke2017automatic}. It heavily uses features from PyTorch such as 
autograd, the rich family of optimizers and the transparent utilization of GPUs. In addition, SimpleITK \cite{lowekamp2013design} is included 
for data input/output to support all standard image file formats. \airlab{} comes along with state-of-the-art
registration components including various image similarity measures, regularization terms and optimizers.
Experimenting with such building blocks or trying out new ideas, for say a regularizer, becomes almost effortless
as gradients are computed automatically. Finally, example implementations
of standard image registration methods are provided such as Optical Flow \cite{horn1981determining}, Demons \cite{thirion1998image} and 
Free Form Deformations \cite{rueckert1999nonrigid}. Deep learning based models are currently not implemented in the AIRLab framework, but we will
integrate them in future releases.
\airlab{} is licensed under the Apache License 2.0 and available on 
GitHub\footnote{\url{https://github.com/airlab-unibas/airlab}}.

In the following, we first provide a brief background about medical image registration followed by 
the description of \airlab{}, its building blocks and its features. Finally, we provide registration
experiments with standard registration methods which are implemented in \airlab{} including performance
analyses and code snippets.

The present draft of this paper roughly introduces \airlab{} and is intended for the presentation at the
8th International Workshop on Biomedical Image Registration in Leiden. A more detailed final version will follow
soon.

\section{Background}
\subsection{Image Registration}
Let ${\mathcal{X}:=\{x_i\}_{i=1}^N}$ be a set of $N$ points arranged in a regular grid which covers the
joint image domain of a moving and fixed image ${\moving,\fixed:\mathcal{X}\rightarrow\R}$. The images map the
$d$-dimensional spatial domain $\mathcal{X}\subset\R^d$ to intensity values. Furthermore, let 
${f:\mathcal{X}\rightarrow\R^d}$ spatially transform the coordinate system of the moving image.
Image registration can be formulated as a regularized minimization problem
\begin{equation}
    f^* = \argmin_{f} \mathcal{S}_\mathcal{X}(\moving\circ f, \fixed) + \lambda\mathcal{R}(f,\mathcal{X}),
    \label{eq:objective}
\end{equation}
where $\mathcal{S}_\mathcal{X}$ is a similarity measure between the transformed moving image and the fixed image
and $\mathcal{R}$ is a regularization term which operates on $f$ on the domain $\mathcal{X}$. The two terms
are balanced by $\lambda$ and $\circ$ is the function composition. 
An example for a similarity measure is
the mean squared error measure for monomodal image registration
\begin{equation}
 \mathcal{S}_{\text{MSE}} := \frac{1}{\vert \mathcal{X} \vert}\sum_{x\in\mathcal{X}} \Big(\moving\big(x+f(x)\big) - \fixed\big(x\big)\Big)^2,
 \label{eq:mse}
\end{equation}
where $\vert\cdot\vert$ is the cardinality of a set. An exemplary regularization term is the diffusion
regularization which favours smooth transformations
\begin{equation}
 \mathcal{R}_{\text{diff}} := \frac{1}{\vert \mathcal{X} \vert}\sum_{x\in\mathcal{X}} \sum_{i=1}^d \Vert\nabla f_i(x)\Vert^2
 \label{eq:diffusion_regularizer}
\end{equation}
where $i$ indexes the space dimension. In the Sections~\ref{sec:features} and \ref{sec:upcoming_features}, the similarity
measures and regularizers which are implemented in \airlab{} are described in more detail.

\paragraph{Transformation}
Transformation models $f$ can be divided in basically four types: \emph{linear}, \emph{non-linear/dense}, \emph{non-linear/interpolating} and \emph{hybrid}. 
Linear transformations, available in \airlab{},
transform each point $x$ with a linear map~$A$
\begin{equation}
 f(x) := A\tilde{x},
\end{equation}
where $A$ is an rotation/translation matrix up to $3$ in 2D and $6$ degrees of freedom in 3D and $x$ stands in homogeneous coordinates $\tilde{x}$.
The class non-linear/parametric transform models consits mainly of two types: interpolating models and dense models.
Non-linear/interpolating transformations
are defined in an interpolating basis on a coarse grid of $n<N$ control points
\begin{equation}
 f(x) := \sum_{i=1}^n c_ik(x_i,x),
 \label{eq:kernel_basis}
\end{equation}
where ${c_i\in\R^d}$ and ${k:\mathcal{X}\times\mathcal{X}\rightarrow\R}$ is the basis function. Common basis functions are the B-spline\cite{rueckert1999nonrigid} or 
Wendland kernel\cite{jud2016sparse} which both are implemented as example basis in \airlab{} (see Section~\ref{sec:features}). The
advantage of non-linear/interpolating transformation models are, that they are computationally efficient. Furthermore, if $k$ is smooth they inherently yield
smooth transformations. 

In non-linear dense transformation models, each point in the image can be transformed individually in $d$ dimensions giving a maximum
flexibility. To still be able to reach reasonable registration results the regularization term is inevitable. 
Hierarchical models can be seen as hybrid interpolating and dense models. Their hierarchical structure enables them to capture large
deformations \cite{jud2018inhomogeneous}. 
Dense transformation models are supported as well by \airlab{} while hybrid are planned.

\paragraph{Optimization}
The similarity measure depends nonlinearly on the moving image $I_{\text{M}}$, which makes an analytical solution to Equation~\ref{eq:objective} intractable. Because in non-linear registration
the number of parameters of $f$ is in the millions, gradient based optimization is usually the only choice to reach a \emph{locally} optimal transformation $f^*$.
Having PyTorch at hand, state-of-the-art optimizers
are available in \airlab{} such as LBFGS \cite{byrd1994representations} and Adam \cite{kingma2014adam}.

\subsection{Image Registration Frameworks}
There are already a considerable amount of medical image registration frameworks available which are valuable
enrichments to the community. Their focus and intensions are diverse, ranging from rich frameworks to specific 
implementations. For an exhaustive list and comparison of such image registration software we refer to \cite{keszei2017survey}.
Gradient free approaches as e.g.~the MRF-based method of \cite{glocker2008dense} are out of scope of the current
implementation of \airlab{}.

The Insight Segmentation and Registration Toolkit (ITK) \cite{yoo2002engineering} is a comprehensive
framework for diverse image processing and analysis task written in C++. It is mainly intended for the use as a library for 
developers who want to implement ready-to-use software. The registration tool Elastix \cite{klein2010elastix} 
is based on ITK and provides a collection of algorithms commonly used in image registration. It can also be used out-of-the-box with the possibility 
of a detailed configuration script. Furthermore, its plug-in architecture
allows to integrate custom parts into the software. The extension SimpleElastix \cite{marstal2016simpleelastix}
offers bindings to other languages such as Python, Java, Ruby and more. Elastix and SimpleElastix are strong if one
needs some flexibility in choosing and combining different registration components for a specific registration task.
SuperElastix is a registration framework, that allows the combination of different existing registration frameworks \cite{Berendsen2016}.
Scalable Image Analysis and Shape Modelling (Scalismo) \cite{bouabene2015scalismo,luthi2017gaussian} is a library mainly for statistical shape modeling written in scala.
It provides also image registration functionality and can be interactively executed similar to SimpleElastix.
Advanced Normalization Tools (ANTs) \cite{avants2011reproducible} is based on ITK as well. It provides a command line tool
including large deformation registration algorithms with standard similarity measures.
The Automated Image Registration software AIR \cite{woods1998automated} is written in C and provides basic registration
functionality for linear and polynomial non-linear image alignment up to the twelfth order.
The Medical Image Registration ToolKit (MIRTK) \cite{rueckert1999nonrigid,schnabel2001generic} is a collection of libraries
and command-line tools for image and point-set registration. Various registration methods based on free form deformations
are provided. Flexible Algorithms for Image Registration (FAIR) \cite{modersitzki2009fair} is a software package
written in MATLAB comprising various similarity measures and regularizers.

None of the mentioned software packages are suited for rapid prototyping in the development of image registration algorithms.
This is mainly because: (I) For the optimization, gradients have to be provided explicitly. For complex transformation models, regularization
terms and similarity measures, their implementation is highly error-prone. (II) For medical images, the computational demand is usually high
and therefore the execution has to be performed on a GPU. The development for GPUs without an appropriate framework is
not trivial. (III) The majority of the frameworks are written in C++. Thus, the development within those frameworks needs good expertise in this language.
Furthermore, the number of code lines required for C++ implementations in these frameworks do not agree with the concept of rapid prototyping.

\section{Autograd Image Registration Laboratory}
\label{sec:airlab}
\airlab{} is a rapid prototyping environment for medical image registration. Its unique characteristics are the automatic differentiation
and the transparent usage of GPUs. It is written in the scripting language Python and heavily uses key functionality of PyTorch \cite{paszke2017automatic}.

The main building blocks constitute:
\begin{itemize}
 \item Automatic differentiation
 \item Similarity measures
 \item Transformation models
 \item Image warping
 \item Regularization terms
 \item Optimizers
\end{itemize}

\subsection{Automatic Symbolic Differentiation}
A key feature of \airlab{} is its automatic \emph{symbolic} differentiation of the objective function. This means, that only the forward function has to be provided by the
developer and the gradient which is required for the optimization is derived through automatic differentiation (AD).
\airlab{} borrows the AD functionality of PyTorch. It is one of the fastest dynamic AD frameworks currently available. Its strong points are:
\begin{itemize}
 \item \emph{Dynamic}: the function which is \emph{symbolically} differentiated is defined by the computations which are run on the variables. Hence, no static graph structure has to be built which fosters rapid prototyping.
 \item \emph{Immediate}: only tensor computations which are necessary for differentiation are recorded
 \item \emph{Core logic}: a low overhead is needed as the AD logic is written in C++ and was carefully tuned 
\end{itemize}
Please cf.~\cite{paszke2017automatic} for more details.

\subsection{Similarity Measures}
\label{sec:features}
We list here the main building blocks required for medical image registration which are provided by \airlab{}.

\begin{itemize}
 \item \textbf{Mean Squared Errors (MSE)}: a simple and fast to compute point-wise measure which is well suited for monomodal image registration
 \begin{equation} 
      \mathcal{S}_{\text{MSE}} := \frac{1}{\vert \mathcal{X} \vert}\sum_{x\in\mathcal{X}} \Big(\moving\big(x+f(x)\big) - \fixed\big(x\big)\Big)^2.
 \end{equation}
 Class name: \texttt{MSE}

 \item \textbf{Normalized Correlation Coefficient (NCC)}: a point-wise measure as $\mathcal{S}_{\text{MSE}}$. It is targeted to image registration tasks, where the intensity relation
 between the moving and the fixed images is linear
 \begin{equation} 
      \mathcal{S}_{\text{NCC}} := \frac{\sum \fixed\cdot (\moving\circ f) - \sum\text{E}(\fixed)\text{E}(\moving\circ f)}{\vert\mathcal{X}\vert\cdot\sum\text{Var}(\fixed)\text{Var}(\moving\circ f)},
 \end{equation}
 where the sums go over the image domain $\mathcal{X}$, $\text{E}$ is the expectation value (or mean) and $\text{Var}$ is the variance of the respective image.\\
 Class name: \texttt{NCC}

 \item \textbf{Local Cross Correlation (LCC)}: is the localized version of $\mathcal{S}_{\text{NCC}}$ where the expectation value and the variance for a given $x$ are computed 
 in a local neighborhood of $x$. In \airlab{} $\mathcal{S}_{\text{LCC}}$ is implemented with efficient convolution operations. Notice that the exact gradient is
 computed using autograd and no gradient approximation is performed in contrast to \cite{cachier20003d}.\\
 Class name: \texttt{LCC}

\item \textbf{Structural Similarity Index Measure (SSIM)}: is a generalization of the LCC similarity measure and was presented by \cite{Wang2004} as an image quality criterion.
The SSIM for two local image patches $a \in \R^d$, $b \in \R^d$  is defined as
\begin{equation}
\text{SSIM}(a, b) = l(a, b)^{\alpha}c(a, b)^{\beta}s(a, b)^{\gamma},
\label{eq:ssim}
\end{equation}
with $\alpha, \beta, \gamma \in [0, 1]$. The SSIM combines three different similarity measures: the luminance ($l$)
\begin{equation}
l(a, b) = \frac{2\mu_a\mu_b + c_1}{\mu_a^2 + \mu_b^2 + c_1},
\label{eq:ssim_lumiation}
\end{equation}
the contrast ($c$)
\begin{equation}
c(a, b) = \frac{2\sigma_a\sigma_b + c_2}{\sigma_a^2 + \sigma_b^2 + c_2},
\label{eq:ssim_contrast}
\end{equation}
and the and structure ($s$)
\begin{equation}
s(a, b) = \frac{\sigma_{ab} + c_3}{\sigma_a\sigma_b + c_3}.
\label{eq:ssim_structure}
\end{equation}
Here, $\mu$ is the mean, $\sigma$ the standard deviation of an image patch, $\sigma_{ab}$ the correlation coefficient, and
$c_1, c_2, c_3 \in \R$ are used to reduce numerical instabilities.
For a complete image the SSIM is defined as 
\begin{equation}
\mathcal{S}_{\text{SSIM}}(\fixed, \moving, f) = \frac{1}{|\mathcal{X}|}\sum_{\substack{x \in \fixed \\ y \in \moving \circ f}}  l[x, y]^{\alpha}c[x, y]^{\beta}s[x, y]^{\gamma}.
\end{equation}
Class name: \texttt{SSIM}

 \item \textbf{Mutual Information (MI)}: was presented as image similarity measure for multimodal image registration by \cite{viola1997alignment,Wells1996}. It is defined as 
 \begin{equation}
\mathcal{S}_{\text{MI}}(\fixed, \moving, f) := H(\fixed) + H(\moving \circ f) - H(\fixed, \moving \circ f),
\end{equation}
where $H(\cdot)$ is the marginal entropy and $H(\cdot,\cdot)$ the joint entropy.
Class name: \texttt{MI}

\item \textbf{Normalized Gradient Fields (NGF)}: is a image similarity measure defined as 
\begin{equation}
\mathcal{S}_{\text{NGF}}(\fixed,I_ M, f) =  \frac{1}{|\mathcal{X}|}\sum_{x \in \mathcal{X}} ||n(\fixed, x) \times n(\moving \circ f, x)||^2, 
\end{equation}
with
\begin{align}
n(I, x)_{\mathcal{E}} &:= \frac{\nabla I(x)}{||\nabla I(x)||_{\mathcal{E}}}, \\
||\nabla I(x)||_{\mathcal{E}} &:= \sqrt{\nabla I(x)^{\text{T}}\nabla I(x) + \mathcal{E}^2},
\end{align}
for multimodal image registration developed by \cite{Haber2006}.

For the estimation of $\mathcal{E}$,  \cite{Haber2006} propose
\begin{equation}
\mathcal{E} = \frac{\eta}{|X|} \sum_{x \in X} |\nabla I(x)|,
\end{equation}
were $\eta$ is the estimated noise level.
Class name: \texttt{NGF}
\end{itemize}

\subsection{Transformation Models}
\airlab{} supports three major types of transformation models: linear/dense, non-linear/interpolating and  dense models (hybrid models are planned).
\subsubsection{Linear/dense}
Currently, \airlab{} supports rigid, similarity and affine transformations for 2D and 3D image data.\\
Class name: \texttt{RigidTransformation}, \newline \texttt{SimilarityTransformation}, \texttt{AffineTransformation}

\subsubsection{Non-linear/interpolating} as mentioned with Equation~\ref{eq:kernel_basis}, non-linear/interpolating models have fewer control points as image points are available. The displacement $f(x)$ for
 a given point $x$ in the image is interpolated from neighboring control points by the respective basis function. 
 In \airlab{}, two exemplary basis functions are implemented:
 \begin{itemize}
  \item B-spline: the standard B-spline kernel, which is used in the Free Form Deformation (FFD) algorithm of \cite{rueckert1999nonrigid}
  \begin{align}
   k_{B_\text{1D}}(x,y) := &
   \begin{cases}
    \frac{2}{3}-\vert r\vert^2 + \frac{\vert r \vert^3}{2}, & 0 \le \vert r\vert < 1\\
    \frac{(2-\vert r\vert)^3}{6}, & 1 \le \vert r\vert < 2\\
    0, & 2 \le \vert r \vert,
   \end{cases}\\
   r = &  x-y .
  \end{align}
  In addition, \airlab{}
  supports B-spline kernels of arbitrary order (first order are used in \cite{vishnevskiy2017isotropic} and third order in the FFD \cite{rueckert1999nonrigid}). An order $p$ is derived by
  convolving the zeroth order B-spline $p+1$ times with it self:
  \begin{align}
   B_0(r) := &  
   \begin{cases}
                 1 & \vert r \vert < \frac{\delta}{2}\\
                 0 & \text{otherwise}
   \end{cases}\\
   B_i := & B_0 * B_{i-1}
  \end{align}
  where $B_3$ corresponds to $k_{B_\text{1D}}$ and $*$ is the convolution.
  The control points have a spacing of $\delta$ which implicitly defines the extent of the kernel.
  With increasing order, the control point support of the kernel is increased by one for each additional order.\\
  Class name: \texttt{BsplineTransformation}
  
  \item Wendland: a family of compact radial basis functions, which is used for image registration in \cite{jud2016sparse,jud2016bilateral}. \airlab{} supports
  a Wendland kernel which is in $C^4$:
  \begin{align}
   k_W(x,y) =& \psi_{3,2}\left(\frac{\Vert x-y\Vert}{\sigma}\right),\\
   \psi_{3,2}(r) =& (1-r)^6_+\frac{3+18r+35r^2}{3},
  \end{align}
  where $a_+ = \max(0,a)$ and $\psi_{3,2}$ is the Wendland function of the second kind and positive definite in $d\le 3$ dimensions.
  The scaling $\sigma$ can also be provided for each space dimension separately to achieve an anisotropic support.\\
  Class name: \texttt{WendlandKernelTransformation}
 \end{itemize}

 For the non-linear/interpolating transformation models, the transposed convolution is applied (cf.~\cite{dumoulin2016guide}) which is available in PyTorch. 
 It is an up-sampling operation
 where the interpolation kernel can be provided. That means in our case, the control points are ``up-sampled'' and interpolated using 
 the basis function of choice.

 \subsubsection{Non-linear/dense} the simpler model is the dense transformation model, where each point in the image can be independently transformed. That means, there are
$nd$ parameters (number of image points times number of space dimensions). To achieve a meaningful transformation, strong regularization is required.

\subsubsection{Image Warping}
To compare the transformed moving image with the fixed image within the similarity measures the coordinate system of the
moving image has to be warped. As it is mostly done in image registration, \airlab{} performs backward warping. That means, 
the transformation is defined on the fixed image domain where the displacement vectors point to the corresponding points in
the moving image. To transform the moving image, it is backward warped into the coordinate system of the fixed image. This
prevents holes occuring in the warped image. 

The warping is performed in normalized coordinates in the interval $[-1,1]^d$. The points which are transformed out of the
fixed image region are identified by checking if $x + f(x)$ falls outside the normalized interval. For illustration please see following snippet:
\begin{lstlisting}[language=Python,commentstyle=\color{gray}]
(...)
displacement = self._grid + displacement

mask = th.zeros_like(self._fixed_image.image, 
		     dtype=th.uint8, device=self._device)
for dim in range(displacement.size()[-1]):
    mask += displacement[..., dim].gt(1) + 
            displacement[..., dim].lt(-1)

mask = mask == 0
(...)
\end{lstlisting}

Because displaced points not necessarily fall 
onto the pixel-grid, interpolation is required. Currently, \airlab{} supports linear interpolation while B-spline interpolation
is planned as up-coming feature. The warping is performed by the grid sampler of PyTorch which utilizes the GPU.

\subsection{Diffeomorphic Transformation}
Diffeomorphic transformations models are often used in medical image registration because of their topology preserving characteristics.
These types of transformations defining a bijective transformation between the fixed image domain and the moving image domain.
First approaches of diffeomorphic image registration were presented in \cite{trouve1995,Christensen1996,Dupuis1998}.
With this the large deformation diffeomorphic metric mapping (LDDMM) was presented by \cite{Beg2005}. 
The LDDMM method possesses a high computational complexity, due to the time dependent velocity used for the calculation of the final transformation.
In order to reduce the computational complexity the usage of a stationary velocity field was presented by \cite{Arsigny2006,Ashburner2007,Hernandez2007}.
The final transformation is then defined as 
\begin{equation}
f = \exp(v),
\label{eq:mat_exp}
\end{equation}
where $\exp(\cdot)$ defines the matrix exponential and $v: \mathcal{X} \to \R^d$ the input vector field.
In this setting the inverse transformation $f^{-1}$ can be obtained by
\begin{equation}
f^{-1} = \exp(v)^{-1} = \exp(-v).
\end{equation}

In \airlab{} diffeomorphic transformation are supported for \emph{all} interpolating and dense transformation models.
We based our implementation in \airlab{} on previous implemented diffeomorphic transformations \cite{abadi2016tensorflow,krebs2018learning}.

\subsection{Different Image Domain Size}
The registration of images with different image domains is a common problem in the field of medical image registration. 
Such problems occur for example if the image modality of both images differ.
We consider two image domains as different, if the extent or the spacing of the fixed and the moving image are different.
Handling different image domain is challenging in a pixel-level environment due to the fact that we directly operate on the pixel level. 
In order to handle images with different image domains, we resample both image to the same pixel spacing and extend the image
size if needed. The computational complexity, which is increased by extending the image size is normally negligible due to the 
highly optimized operations used on the GPU.

\subsection{Regularization}
There are three different types of regularization terms in \airlab{}. (I) Regularizers on the 
displacement field $f$ commonly used in FFD registration, (II) regularizers on the parameters of
$f$ elaborated in \cite{jud2016sparse,jud2016bilateral,vishnevskiy2017isotropic} and (III) the Demons regularizers
which regularize the displacement field $f$ by filtering it in each iteration. Note that Demons regularizers
are not differentiated, because in Demons approaches the optimization is an iteration scheme where
the image forces (gradient of similarity measure) are evaluated to update the current displacement field
and alternatingly the displacement field is regularized using filtering.
We first list the regularization terms which operate on the displacement field $f$.
\begin{itemize}
 \item Diffusion: a regularizer which penalizes changes in the transformation $f$
    \begin{equation}
    \mathcal{R}_{\text{diff}} := \frac{1}{\vert \mathcal{X} \vert}\sum_{x\in\mathcal{X}} \sum_{i=1}^d \big\Vert\nabla f_i(x)\big\Vert^2_2.
    \end{equation}
    Class name: \texttt{DiffusionRegulariser}
 \item Anisotropic Total Variation: a regularizer which favours piece-wise smooth transformations $f$
    \begin{equation}
    \mathcal{R}_{\text{anisoTV}} := \frac{1}{\vert \mathcal{X} \vert}\sum_{x\in\mathcal{X}} \sum_{i=1}^d \big\vert\nabla f_i(x)\big\vert.
    \end{equation}
    It is anisotropic which means
    its influence is aligned to the coordinate axes.\\
    Class name: \texttt{TVRegulariser}
 \item Isotropic Total Variation: the isotropic version of the anisotropic regularizer
    \begin{equation}
    \mathcal{R}_{\text{isoTV}} := \frac{1}{\vert \mathcal{X} \vert}\sum_{x\in\mathcal{X}} \big\Vert\nabla f(x)\big\Vert_2.
    \end{equation}
    Both TV regularizers are not differentiable, therefore, the subgradient of zero is taken at zero.\\
    Class name: \texttt{IsotropicTVRegulariser}
 \item Sparsity: a regularizer which penalizes non-zero parameters
  \begin{equation}
  \mathcal{R}_{\text{sparse}} := \frac{1}{\vert \mathcal{X} \vert}\sum_{x\in\mathcal{X}} \big\Vert f(x)\big\Vert_1.
  \end{equation}
  Class name: \texttt{SparsityRegulariser}
\end{itemize}

\subsubsection{Regularizers on Parameters}
The listed regularization terms are also available for regularizing the parameters of $f$. The parameters which should be regularized
are passed to the regularizer as an array, a name and a weighting. In this way, one can individually weight subsets of parameters,
belonging for example to different hierarchical levels, cf.~the following example:
\begin{lstlisting}[language=Python,commentstyle=\color{gray}]
(...)
 reg_param = paramRegulariser.L1Regulariser(
		    "trans_parameter", 
		    weight=weight_parameter[level])
		
 registration.set_regulariser_parameter([reg_param])
(...)
\end{lstlisting}

\subsubsection{Demons Regularizers}
Currently, there are two Demons regularizers available in \airlab{}:
\begin{itemize}
 \item Kernel: an arbitrary convolution kernel for filtering the displacement field. An example is the Gaussian
 kernel which is used originally in the Demons algorithm \cite{thirion1998image}.\\
 Class name: \texttt{GaussianRegulariser}
 \item Graph Diffusion: the diffusion is performed by spectral graph diffusion. The graph can be utilized in order to handle 
 the sliding organ problem. In this case, the graph is built during the optimization as proposed by \cite{robin2018adaptive}.\\
 Class name: \texttt{GraphDiffusionRegulariser}
\end{itemize}

\subsection{Optimizers}
\airlab{} includes a rich family of optimizers which are available in PyTorch including LBFGS, ASGD and Adam. They are tailored to optimize functions with a high number of 
parameters and thus are well suited for non-linear image registration objectives. We refer to \cite{ruder2016overview} for a detailed overview
of first order gradient based optimizers. As PyTorch also supports \emph{no-grad} computations, iteration schemes as used in the Demons algorithm
are also supported. The following snippet is an example usage of \emph{no-grad} taken from the Demons regularizer.
\begin{lstlisting}[language=Python,commentstyle=\color{gray}]
(...)
def regularise(self, data):
    for parameter in data:
        # no gradient calculation for the 
        # demons regularisation
        with th.no_grad():
            self._regulariser(parameter)
(...)
\end{lstlisting}

\begin{table*}[]
\begin{center} 
  \caption{Execution statistics of 2D images with different sizes (pixels) in seconds.}
  \label{table:performance_table}
    \begin{tabularx}{0.8\textwidth}{ cc|cccccccc}
\hline
     Experiment & Hardware & 64 & 128 & 256 & 512 & 1024 & 2048 & 4096 \\
     \hline
    Dense + $\mathcal{R}_{\text{diff}}$  + $\mathcal{S}_{\text{MSE}}$ &  CPU & 2.29 & 3.29 & 5.57 & 14.11 & 46.44 & 187.68 & 832.77\\
    Dense \& Diffeomorph + $\mathcal{R}_{\text{diff}}$  + $\mathcal{S}_{\text{MSE}}$ &  CPU & 4.52	& 9.75 & 25.81 & 81.55 & 309.06 & 1367.49 & 5834.58\\
    Dense + $\mathcal{R}_{\text{diff}}$  + $\mathcal{S}_{\text{MSE}}$ &  GPU GTX 1080 & 4.41 & 4.39 & 4.39 & 4.29 & 4.63 & 9.21 & 30.91\\
    Dense \& Diffeomorph + $\mathcal{R}_{\text{diff}}$  + $\mathcal{S}_{\text{MSE}}$ &  GPU GTX 1080 & 7.36 & 7.32 & 6.59 & 6.61 & 9.14 & 24.34 & 89.85\\ 
    \hline
    \end{tabularx}
\end{center}    
\end{table*}
\subsection{Registration Evaluation}
Performance evaluation of the developed registration method is essential. Due to the fact that ground truth transformations are highly difficult to obtain especially for 
medical images other performance measures are used. Over the last years the evaluation of the registration is performed on a selected set of corresponding landmarks.
Several datasets have been provided to evaluate registration algorithm. 
Here, the POPI \cite{Vandemeulebroucke2011} or the DirLab \cite{castillo2009framework} dataset are mostly used in the past. 
These datasets contain 3D CT images of the upper thorax and corresponding landmarks.
However, normally the developer is responsibly  to implement the necessary functions for the pre-processing of the image data or the evaluation of the landmarks.
In \airlab{}, we provide an evaluation pipeline for download, pre-processing, and 
evaluation. This means in detail, that the developer can plug in the new registration algorithm and \airlab{} take care of the complete evaluation process. For the comparison of
corresponding landmarks, we use the mean square distance of the landmark positions. The included automatic evaluation pipeline makes it very simple to compare various 
versions of the registration algorithm during the development. To the best of our knowledge, \airlab{} is the first registration framework that provide such 
an automatic evaluation pipline. The automatic evaluation is done for the POPI dataset \cite{Vandemeulebroucke2011}.

\subsection{Upcoming Features}
\label{sec:upcoming_features}
In this section, we list the features which did not make it into the present version, which however are planned for integration into \airlab{} soon.
\begin{itemize}
   \item Interpolation: B-spline interpolation for the image warping.
   \item Registration: Registration of images with more than one color channel, e.g., RGB images.
   \item Registration: Extending \airlab{} for learning-based registration.
\end{itemize}

\section{Experiments}
In this section, we provide image registration examples. We have implemented two classic registration algorithms within \airlab{}
and show their qualitative performance on synthetic examples and on a DirLab dataset \cite{castillo2009framework}. Quantitative analyses will follow
in the final version of this paper.
\begin{figure}[H]
\centering
 \subfloat[Fixed/Moving]{%
 \raisebox{0mm}{
    \includegraphics[angle=0,width=0.2\textwidth]{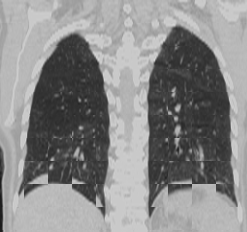}
    }
 }
\subfloat[Fixed/Warped]{%
\raisebox{0mm}{
   \includegraphics[angle=0,width=0.2\textwidth]{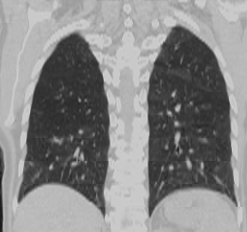}
   }
}\\
\subfloat[Transformation]{%
\raisebox{0mm}{
   \includegraphics[angle=0,width=0.2\textwidth]{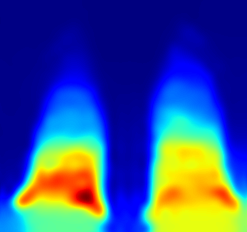}
   }
}
\caption{\emph{FFD} registration result. (a) Fixed image and moving image as checkerboard, (b) fixed image and warped moving image as 
checkerboard and (c) final transformation visualized as the magnitudes of the displacements.}
\label{fig:ffd_example}
\end{figure}

\subsection{Image Registration Algorithms}
\begin{figure*}[t]
\centering
 \subfloat[Fixed]{%
 \raisebox{0mm}{
    \includegraphics[width=0.2\textwidth]{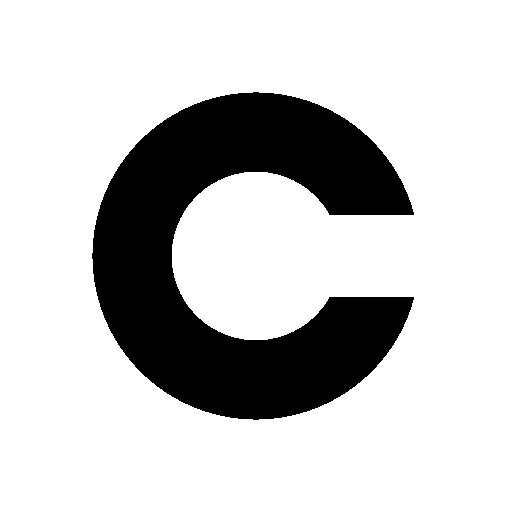}
    }
 }
\subfloat[Moving]{%
\raisebox{0mm}{
   \includegraphics[width=0.2\textwidth]{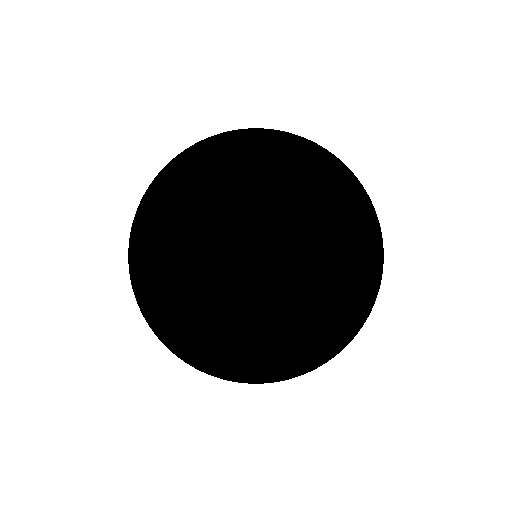}
   }
}
\subfloat[Moving for warping]{%
\raisebox{0mm}{
   \includegraphics[width=0.2\textwidth]{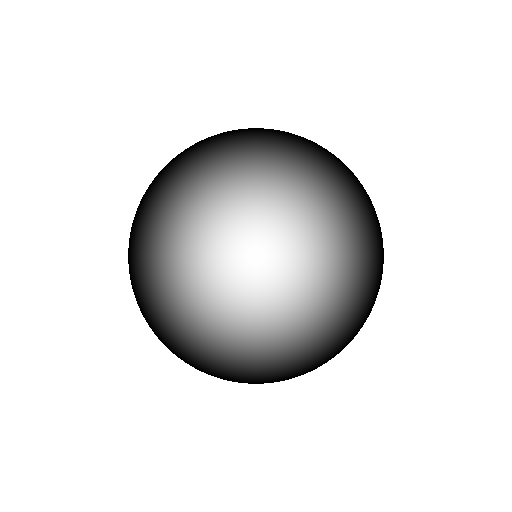}
   }
}\\
\subfloat[Warped (Demons)]{%
\raisebox{0mm}{
   \includegraphics[width=0.2\textwidth]{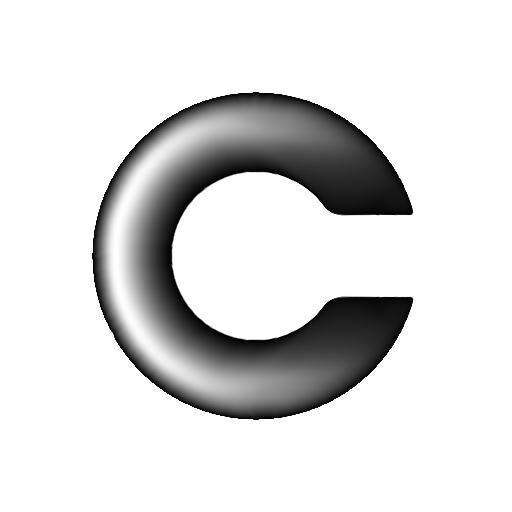}
   }
}
\subfloat[Transformation (Demons)]{%
\raisebox{0mm}{
   \includegraphics[width=0.2\textwidth]{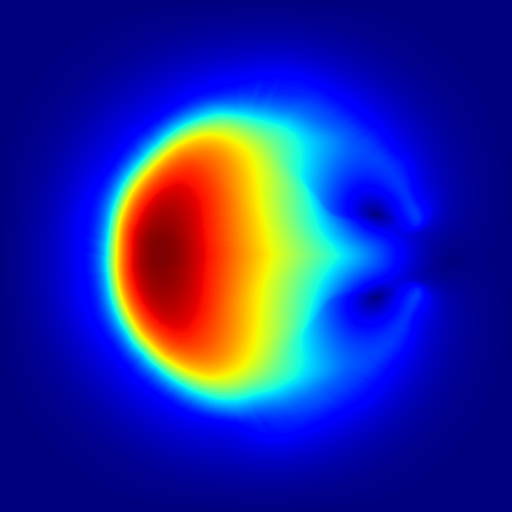}
   }
}
\subfloat[Inverse Transformation \newline  (Demons)]{%
\raisebox{0mm}{
   \includegraphics[width=0.2\textwidth]{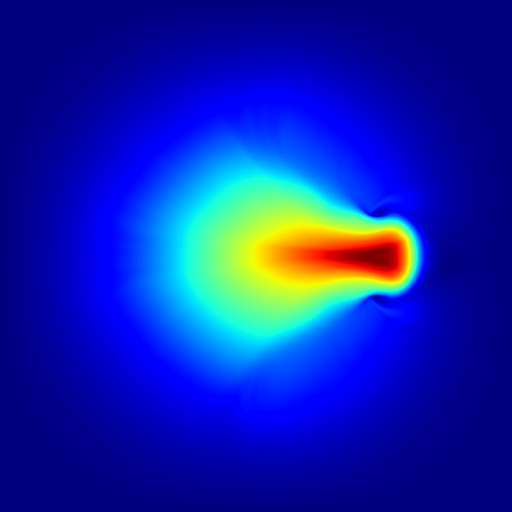}
   }
}
\subfloat[Reconstration of the moving image from (d) with (f) (Demons)]{%
\raisebox{0mm}{
   \includegraphics[width=0.2\textwidth]{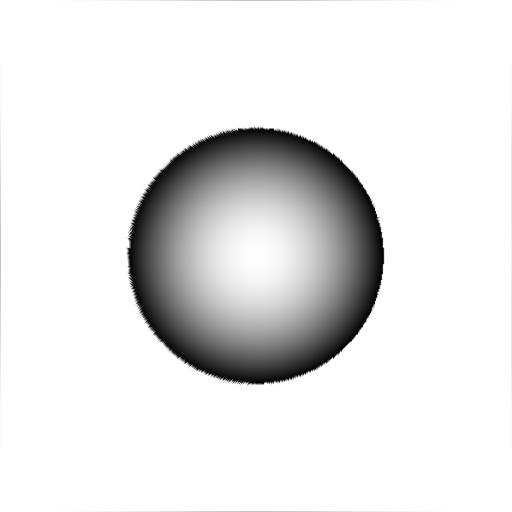}
   }
}\\
\subfloat[Warped (B-spline)]{%
\raisebox{0mm}{
   \includegraphics[width=0.2\textwidth]{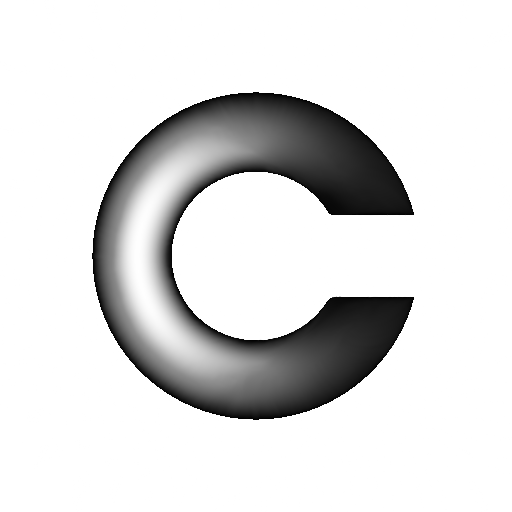}
   }
}
\subfloat[Transformation (B-spline)]{%
\raisebox{0mm}{
   \includegraphics[width=0.2\textwidth]{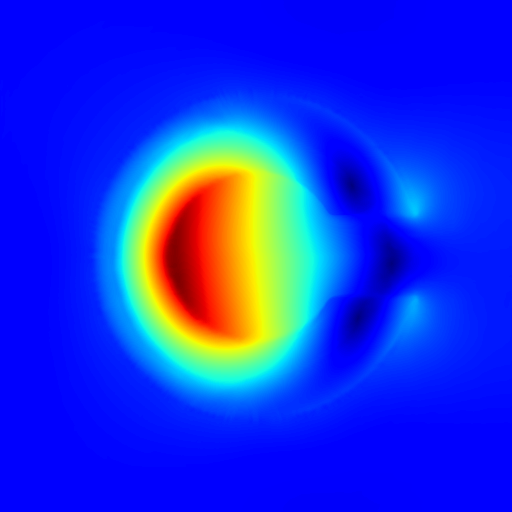}
   }
}
\subfloat[Inverse Transformation \newline (B-spline)]{%
\raisebox{0mm}{
   \includegraphics[width=0.2\textwidth]{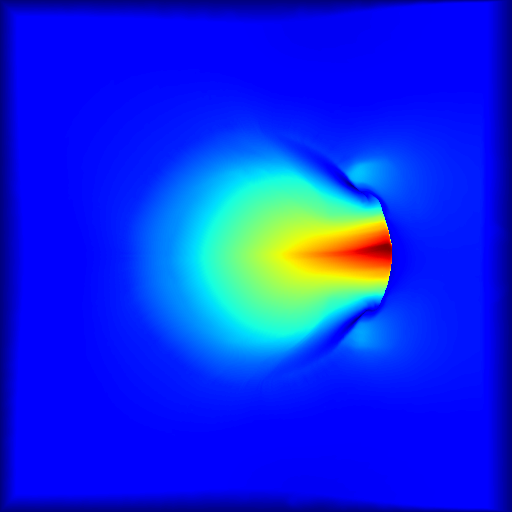}
   }
}
\subfloat[Reconstration of the moving image from (h) with (j) (B-spline)]{%
\raisebox{0mm}{
   \includegraphics[width=0.2\textwidth]{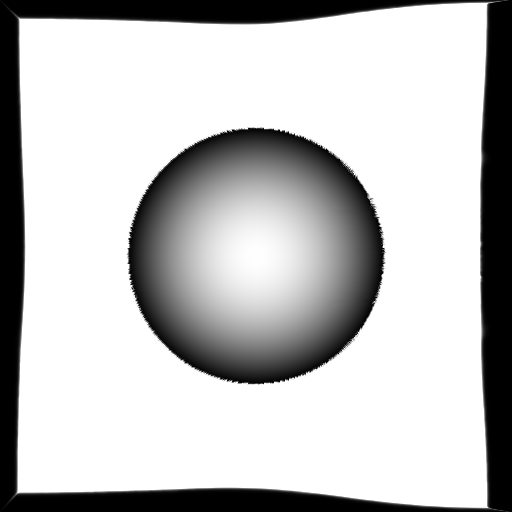}
   }
}

\caption{(a) Fixed C image, (b) moving circle image and (c) shaded circle image. Results for the diffeomorphic demons((d)-(g)): (d) warped shaded circle, (e)
final transformation and (f) inverse transformation visualized as the magnitudes of the displacements and (g) reconstruction of the 
moving image with the inverse transformation. ((h)-(k)) shows the result for the diffeomorphic B-spline registration method.}
\label{fig:demons_example}
\end{figure*}

The following algorithms have been implemented:
\begin{itemize}
 \item \emph{Rigid}: a simple objective with a similarity transformation has been set up, where the $\mathcal{S}_{\text{MSE}}$ similarity metric has been optimized with Adam.
               Two \airlab{} images have been registered, where the moving image has been rotated, translated, and scaled. In Figure~\ref{fig:rigid_example}, the registration result is depicted.
 \item \emph{FFD}: the Free Form Deformations algorithm \cite{rueckert1999nonrigid} was implemented. As in the original paper, a third order B-spline kernel 
 has been used for the parametric transformation model. Furthermore, the $\mathcal{S}_{\text{NCC}}$ similarity measure with the $\mathcal{R}_{\text{anisoTV}}$ regularizer
 on the displacement field have been applied. The overall objective has been optimized with Adam. For the experiment, an image pair of the DirLab\cite{castillo2009framework} has been registered. To illustrate the result,
in Figure~\ref{fig:ffd_example}, a slice through the volume is visualized. A multi-resolution strategy has been implemented performing $\{300,200,50\}$ iterations for the \emph{FFD} algorithm.
The detailed parameter configuration can be found in the source-code.
 \item \emph{Diffeomorphic}: the Demons algorithm \cite{thirion1998image} was implemented using the $\mathcal{S}_{\text{MSE}}$ similarity measure with the Gaussian Demons regularizer.
 Furthermore, we used the diffeomorphic option of the transformation. The Diffeomorphic Demons algorithm has been applied to the circle and C example. 
 For better illustration, see Figure~\ref{fig:demons_example} (d)-(g), a shaded circle
 has been warped with the final transformation. In addition, we applied a diffeomorphic B-spline registration to the circle and C example. As similarity measure also the  $\mathcal{S}_{\text{MSE}}$ was used. 
 The results are shown in Figure~\ref{fig:demons_example} (h)-(k), a shaded circle has been warped with the final transformation. 
\end{itemize}

The following snippet illustrates how to setup a registration algorithm in \airlab{} with the \emph{Rigid} registration example:

\begin{lstlisting}[language=Python,commentstyle=\color{gray}]
(...)
# all imports
registration = PairwiseRegistration(dtype=dtype, 
				    device=device)

# choose the rigid transformation model
transformation = SimilarityTransformation(moving_image,
                                          opt_cm=False)

# initialize the translation with the center of mass
  of the fixed image
transformation.init_translation(fixed_image)

registration.set_transformation(transformation)

# choose the Mean Squared Error as image loss
image_loss = MSELoss(fixed_image, moving_image)
registration.set_image_loss([image_loss])

# choose the Adam optimizer to minimize the objective
optimizer = th.optim.Adam(
		transformation.parameters(), lr=0.01)
registration.set_optimizer(optimizer)
registration.set_number_of_iterations(1000)

# start the registration
registration.start()

# warp the moving image with the final transformation result
displacement = transformation.get_displacement()
warped_image = warp_image(moving_image, displacement)
(...)
\end{lstlisting}



\subsection{Performance Analysis}    
All experiments have been conducted using an NVIDIA GeForce GTX 1080 GPU. 
We evaluate the performance of \airlab{} for different image sizes and different computation hardware. Furthermore, we evaluated the computaional effort for the diffeomorphic transformation
compared to the non-diffeomorphic transformation models.
The performance comparism of CPU and GPU is listed in Table~\ref{table:performance_table}.

Because the \texttt{GaussianRegulariser} is not differentiated, there is less computational time spent by autograd for the \emph{Demons} example.

\section{Conclusion}
We have introduced \airlab{}, an environment for rapid prototyping and reproduction of medical image registration algorithms. It is
written in the scripting language Python and heavily uses functionality of PyTorch. The unique feature compared to existing image registration software is the automatic 
differentiation which fosters rapid prototyping. \airlab{} is freely available under the Apache License 2.0 and accessible 
on GitHub: \url{https://github.com/airlab-unibas/airlab}.

With \airlab{}, we hope that we can make a valuable contribution to the medical image registration community, 
and we are looking forward to see researchers and developers which activly use \airlab{} in their work. Finally, we encourage them also to contribute to future
developments of \airlab{}.

\section*{Acknowledgments}
The authors thank Alina Giger, Reinhard Wendler, and Simon Pezold for their great support. 

\bibliographystyle{plain}
\bibliography{bibliography.bib}


\end{document}